\documentclass[runningheads]{llncs}

\usepackage{accv}

\usepackage{accvabbrv}

\usepackage{graphicx}
\usepackage{booktabs}
\usepackage{mathtools}
\usepackage{wrapfig}
\usepackage[accsupp]{axessibility}  
\usepackage{graphicx}
\usepackage{amsmath,amssymb} 
\usepackage{color}
\usepackage{booktabs}
\usepackage{array}
\usepackage{bbm}
\usepackage{booktabs,multirow,array,caption}
\newcolumntype{P}[1]{>{\centering\arraybackslash}p{#1}}
\newcolumntype{L}[1]{>{\arraybackslash}p{#1}}
\newcolumntype{M}[1]{>{\centering\arraybackslash}m{#1}}
\usepackage{dsfont}
\usepackage{graphicx,wrapfig,lipsum}
\usepackage[misc]{ifsym}

\usepackage[pagebackref,breaklinks,colorlinks,citecolor=accvblue]{hyperref}

\usepackage{orcidlink}

\begin{document}

\title{MECFormer: Multi-task Whole Slide Image Classification with Expert Consultation Network} 

\titlerunning{MECFormer}

\author{Doanh C. Bui\orcidlink{0000-0003-1310-5808} \and
Jin Tae Kwak\orcidlink{0000-0003-0287-4097}\textsuperscript{(\Letter)}}

\authorrunning{Doanh C. Bui et al.}

\institute{School of Electrical Engineering, Korea University, Seoul, Republic of Korea
\email{\{doanhbc,jkwak\}@korea.ac.kr}}

\maketitle

\begin{abstract}
Whole slide image (WSI) classification is a crucial problem for cancer diagnostics in clinics and hospitals. A WSI, acquired at gigapixel size, is commonly tiled into patches and processed by multiple-instance learning (MIL) models. Previous MIL-based models designed for this problem have only been evaluated on individual tasks for specific organs, and the ability to handle multiple tasks within a single model has not been investigated. In this study, we propose MECFormer, a generative Transformer-based model designed to handle multiple tasks within one model. To leverage the power of learning multiple tasks simultaneously and to enhance the model's effectiveness in focusing on each individual task, we introduce an Expert Consultation Network, a projection layer placed at the beginning of the Transformer-based model. Additionally, to enable flexible classification, autoregressive decoding is incorporated by a language decoder for WSI classification. Through extensive experiments on five datasets involving four different organs, one cancer classification task, and four cancer subtyping tasks, MECFormer demonstrates superior performance compared to individual state-of-the-art multiple-instance learning models.
\end{abstract}

\section{Introduction}
\label{sec:intro}

In computational pathology, whole slide images (WSIs) have been extensively studied and investigated to provide a comprehensive understanding of tissues and diseases inside the human body and to help pathologists make accurate and reliable diagnoses in clinics \cite{li2022comprehensive}. Due to the gigapixel resolution of WSIs, recent studies have mainly focused on the efficient and effective processing of WSIs, which led to the development of various multiple instance learning (MIL) methods. These MIL-based methods often include two-stage procedures: 1) tiling the WSI into a bag of patches and extracting their features using an off-the-shelf feature extractor; 2) learning a classifier to conduct slide-level classification. There are two primary research directions in MIL-based methods: instance-based and bag embedding-based models. Instance-based models \cite{hou2016patch,campanella2019clinical} provide patch-level predictions, which are then aggregated to make the final prediction. Bag embedding-based models \cite{ilse2018attention,campanella2019clinical,shao2021transmil,lu2021data,zhang2022dtfd}, utilize aggregators to combine all patch features into a bag embedding, and then a bag classifier provides the final prediction. 
There are numerous tasks in pathology, many of which are interrelated to each other. However, most MIL-based models have been tailored and evaluated on an individual task basis under the single-task learning paradigm. This overlooks the potential to learn and leverage knowledge from multiple related tasks to resolve differing tasks, which is similar to the learning and adaptation process of human experts. 
One may develop a single generalizable model that can handle multiple tasks, but it is challenging for a single model to learn knowledge from several tasks simultaneously \cite{guo2020learning}. 


\begin{figure}[http]
\centerline{\includegraphics[width=8cm]{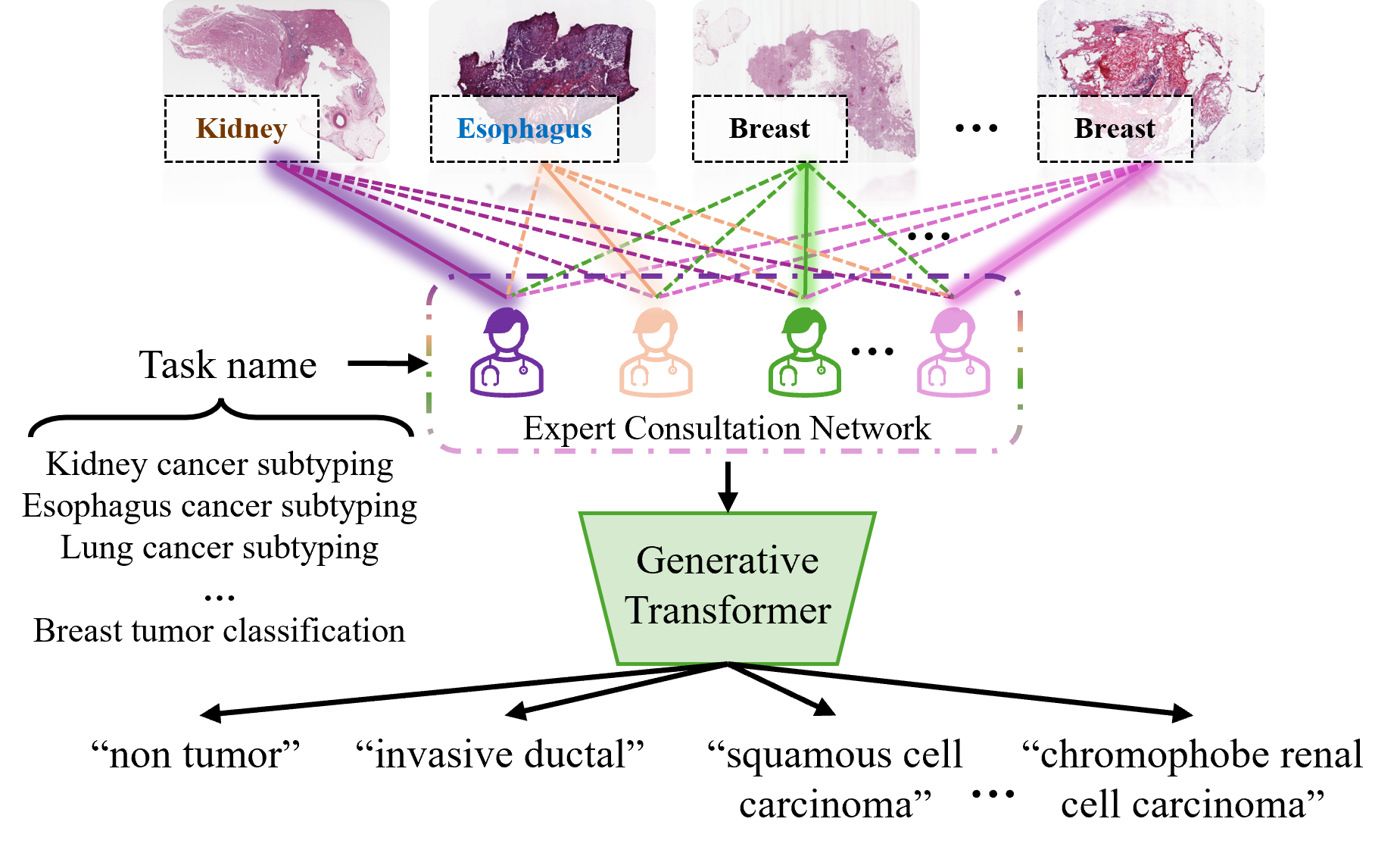}}
\caption{Illustration of the multi-task process of MECFormer. All experts observe and process an input WSI as the corresponding expert to the target task makes the most significant contribution. A generative transformer-based model receives information from all experts and produces the predicted diagnostic term.}
\label{fig:thumbnail}
\end{figure}

To address the above mentioned issues, recent efforts have developed Mixture-of-Experts (MoE) techniques \cite{ye2023taskexpert, chen2023adamv}, which aim to align features with the embedding space of the target task using multiple experts defined as learnable projection modules. These projections are aggregated by a weighting mechanism. MoE techniques have been applied to semantic segmentation and depth estimation tasks \cite{chen2023adamv,ye2023taskexpert}; however, its utility and usefulness in WSI analysis have not been investigated yet. Moreover, most MoE methods adopt a single classification head per task that needs to be separately trained. Furthermore, a language decoder has been developed for pathology image analysis, showing its promising ability for report generation \cite{zhang2020evaluating,tsuneki2022inference} and multi-class cancer classification \cite{gpc} on patch-level pathology images. To the best of our knowledge, no prior work has adopted the language decoder for WSI analysis.

Herein, we propose an approach for \textbf{M}ulti-task WSI classification using an \textbf{E}xpert \textbf{C}onsultation Network (\texttt{ECN}) and a single Trans\textbf{Former}-based architecture, designated as MECFormer, that permits the ability to handle WSIs for various tasks. Figure \ref{fig:thumbnail} depicts the multi-task process of MECFormer. In brief, we design MECFormer for multi-task WSI classification as a generative Transformer-based model. 
Given a WSI, MECFormer builds a bag of patches (or embeddings) and processes it using a Transformer-based encoder.
In the encoder, we devise the first projection layer, i.e., \texttt{ECN}, to project the input embeddings into the model space by combining the knowledge of multiple experts via routering, task assignment, and knowledge aggregation under the guidance of the task indicator. 
Using the task indicator, \texttt{ECN} effectively gathers knowledge across multiple tasks in a controllable manner while highlighting patterns for the target task, leading to effective embeddings.
Then, a Transformer-based decoder takes the knowledge aggregated embeddings along with a conditional textual input to produce the output sequence, i.e., a predicted diagnostic term, in an auto-regressive manner. 
We note that a single Transformer-based decoder is shared among multiple tasks. Meanwhile other MoE models adopt task-specific classification heads. Our implementation is available at \footnote{\url{https://github.com/QuIIL/MECFormer}}.

In summary, our contributions can be listed as below:

\begin{itemize}
\item We propose MECFormer to handle multiple well-known tasks for WSI analysis using a single unified Transformer-based model.
\item We introduce \texttt{ECN} to efficiently and effectively learn and aggregate knowledge from multiple experts on the target task in a controllable manner.
\item We adopt a generative Transformer-based decoder in MECFormer, based up cross-attention between textual and visual embeddings, to enable unified and flexible classification for multiple WSI classification tasks.
\item MECFormer achieves superior performance to other state-of-the-art task-specific MIL models on five datasets/tasks, including CAMELYON16, TCGA-BRCA, TCGA-ESCA, TCGA-NSCLC, and TCGA-RCC.
\end{itemize}

\section{Related Work}
\subsection{Multiple Instance Learning Models for WSI analysis} 
Recent studies of MIL-based models commonly adopt bag embedding-based approaches. These are often built upon an attention mechanism. For example, AB-MIL \cite{ilse2018attention} learns to generate a weight for each patch using the attention mechanism and then aggregates all the patch embeddings using a weighted average. CLAM \cite{lu2021data} introduces an auxiliary task that clusters the most-attended patch embeddings as positive patches and the least-attended patch embeddings as negative patches to constrain and refine the embedding space. TransMIL \cite{shao2021transmil} builds a Transformer-based architecture that approximates self-attention for computational efficiency and adopts positional encoding to retain the spatial information of the cropped patches by reshaping them into a 2-dimensional space and applying multiple learnable convolutions. DTFD-MIL \cite{zhang2022dtfd} is designed in a two-stage manner, consisting of sub-MIL and global MIL branches, to handle over-fitting problems due to the limited number of WSIs. It randomly creates and aggregates multiple sub-bags using a sub-MIL branch, producing sub-bag embeddings. A global MIL branch makes the final prediction based on the sub-bag embeddings.

\subsection{Mixture-of-Experts Techniques for Multi-task Learning}

MoE methods have recently been developed to handle multiple tasks in a single unified model. For instance, \cite{ye2023taskexpert} proposes decomposing the feature maps into a set of multiple representative features that are processed by an expert as an MLP. A gating mechanism is devised to determine the contribution of each feature, which is used to aggregate all representative features via a weighted sum. A feature memory is adopted to capture long-range dependencies of task-specific representation. \cite{chen2023adamv} proposes to use a sparse MoE layer in the Vision Transformer (ViT) \cite{dosovitskiy2020image}, in which a router network is introduced for each task, along with multiple experts as MLPs. Each router network determines how many experts are employed and how they contribute to the task-specific representation. Assuming that only pre-trained models from multiple tasks are accessible, \cite{tang2024merging} utilizes task vectors \cite{ilharco2022editing} and a learnable router to generate weights for knowledge retrieval. These weights are combined with a base model, such as CLIP-ViT-B/32 \cite{radford2021learning}, to produce features better aligned with each task.

\subsection{Language Models for Pathology Analysis} 

Some research efforts have been made to apply and exploit language models for pathology image analysis. For instance, Zhang et al. \cite{zhang2020evaluating} builds a network using ResNet18 to extract visual features and long short-term memory (LSTM) to generate histopathology captions. Tsuneki et al. \cite{tsuneki2022inference} proposes to utilize EfficientNet as a visual encoder and a language decoder based on a recurrent neural network to generate reports for pathology images from gastric adenocarcinoma specimens. In an attempt to deal with multi-task pathology analysis, Nguyen et al. \cite{gpc} proposes GPC for generative pathology classification, which employs a convolutional neural network as a visual encoder and a pre-trained language decoder to predict cancer categories. For slide-level classification, Bryan et al. \cite{guevara2023caption} investigates the possibility of using a language decoder on WSIs. Using a dataset with image-caption pairs, the decoder is designed to learn the visual-text correspondence and to produce captions. Pairs of images and generated captions are then manually classified into a diagnostic category for evaluation.

\section{Methodology}


\begin{figure}[http]
\centerline{\includegraphics[width=11cm]{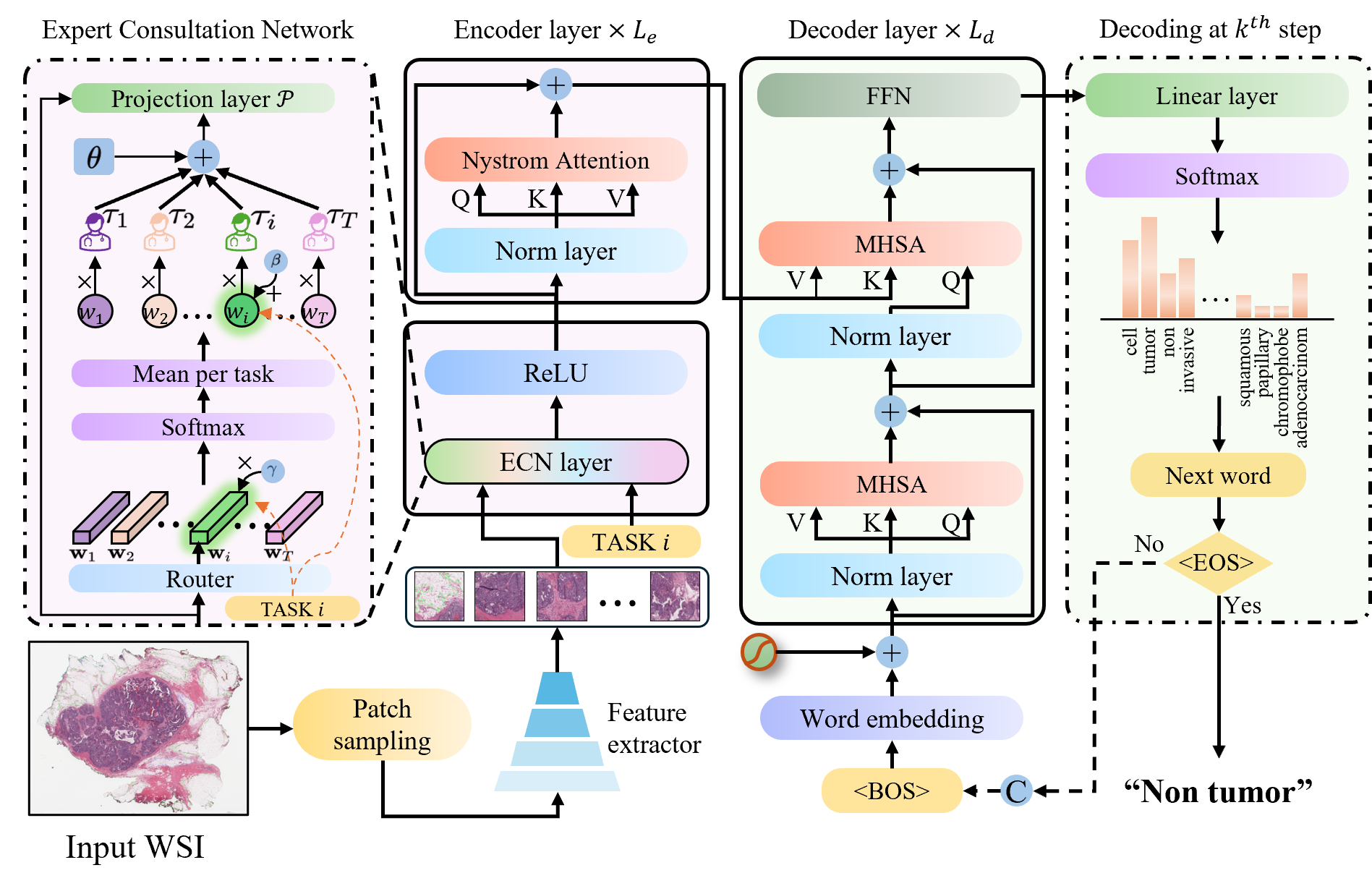}}
\caption{Overview of MECFormer. MECFormer is designed as a Transformer-based generative model in an encoder-decoder manner. \texttt{ECN} layer is placed at the beginning of the encoder and aware of the target task along with the input bag of patch features.}
\label{fig:overview}
\end{figure}

\subsection{Problem Definition} In this study, we aim to develop a Transformer-based model $\mathcal{F}(\cdot)$, MECFormer, which has the ability to handle multiple tasks for WSI classification. Given a WSI $\mathcal{X}$ and a task indicator $\mathcal{T}$, $\mathcal{F}(\cdot)$ produces the output that is tailored to a task of interest $\hat{\mathcal{Y}} = \mathcal{F}(\mathcal{X}, \mathcal{T})$ where $\hat{\mathcal{Y}}$ is the text output. There are $T$ distinct tasks, each with its corresponding training dataset $\mathcal{D}^{train}_t$. These $T$ datasets are merged to form $\mathcal{D}^{train} = \{\mathcal{D}^{train}_t\}_{t=1}^{T}$, which is used to train $\mathcal{F}(\cdot)$. Then, $\mathcal{F}(\cdot)$ is evaluated on the testing set of each $t^{th}$ task, i.e., $\mathcal{D}^{test}_t$. 

\subsection{MECFormer}

Figure \ref{fig:overview} illustrates the overview of MECFormer. MECFormer involves a visual encoder $\mathcal{E}(\cdot)$ and a language decoder $\mathcal{D}(\cdot)$. The role of $\mathcal{E}(\cdot)$ is to encode the bag of patch features as exploring intra-relationship between patches in an efficient and effective manner with Nystr\"om Attention. The bag of patch features are obtained by WSI pre-processing, which is described below. $\mathcal{D}(\cdot)$ learns to align textual and visual embeddings from the encoder, and then performs auto-regressive decoding to predict a diagnostic term. 

The overall procedure of MECFormer can be summarized as follows: 1) Given an input WSI $\mathcal{X}$ and a task indicator $\mathcal{T}$, $\mathcal{X}$ undergoes WSI pre-processing, producing a bag of patch features $\mathbf{x} \in \mathbb{R}^{N \times d_f}$ where $N$ denotes the number patches and $d_f$ refers to the feature dimension; 2) $\mathbf{x}$ is fed into $\mathcal{E}(\cdot)$ where $\mathbf{x}$ is embedded into the model space by an $\texttt{ECN}$ projection layer $\mathcal{P}_{\texttt{ECN}}(\cdot)$ and processed by a series of encoding layers to produce visual token embeddings; 3) $\mathcal{D}(\cdot)$ conducts the decoding process, i.e., generating the next word, over an unknown number of $K$ steps where each next word is conditioned on the previous word embeddings and the visual token embeddings from $\mathcal{E}(\cdot)$. The procedure can be given by:

\begin{equation}
\hat{y}^{(k+1)} = \mathcal{D}\Big(\mathcal{E}\big(\mathcal{P}_{\texttt{ECN}}(\mathbf{x}, \mathcal{T})\big)\Big\vert\hat{y}^{(k)}\Big),
\label{eq:1}
\end{equation}

\noindent where $\hat{y}^{(k)}$ is the $k^{th}$ prediction (or word). $\hat{y}^{(0)} = \texttt{<BOS>}$, which is begin-of-sentence token to start the decoding. When $\hat{y}^{(k+1)} = \texttt{<EOS>}$ is obtained, the decoding process terminates.

\subsection{WSI Pre-processing} 

Given a WSI $\mathcal{X}$, a patch sampling procedure is adopted to tile $\mathcal{X}$ into a bag of patches. In this process, we segment tissue regions on $\mathcal{X}$ and crop patches from the segmented regions to form a bag of patches. Each patch in the bag undergoes an off-the-shelf pre-trained feature extractor $\mathcal{G}(\cdot)$, producing a bag of patch features $\mathbf{x} = \{x_i\}^{N}_{i=1}$ where $x_i \in \mathbb{R}^{d_f}$ and $d_f$ is determined by $\mathcal{G}(\cdot)$. In this study, we adopt two off-the-shelf pre-trained feature extractors: CTransPath \cite{wang2022transformer} and UNI \cite{chen2024towards}, which are pre-trained models that are trained on large-scale pathology image datasets via self-supervised learning. 


\subsection{Expert Consultation Network (\texttt{ECN})}

To enhance the learning capability for multiple tasks, we devise an effective projection layer, $\mathcal{P}_{\texttt{ECN}}(\cdot)$, whose primary role is to project the input patch embeddings into the most suitable model space using the aggregated knowledge of multiple experts. 
Suppose that there exist multiple experts $\boldsymbol{\tau} = \{\tau_i\}_{i=1}^{T}$ that are designed to acquire knowledge for the corresponding tasks, and the common knowledge $\theta_p$. During training, we learn how to obtain the knowledge of each expert and aggregate the knowledge in three stages: 1) \textit{Preliminary Consultation}, 2) \textit{Expert Assignment}, and 3) \textit{Expert Consultation}. 
In Preliminary Consultation, a router $\mathcal{R}(\cdot)$ is designed to initialize $T$ weight vectors that specify how to combine knowledge from $T$ experts.
In Expert Assignment, each expert $\tau_i$ is associated with its corresponding task by adjusting the $T$ weight vectors. 
Finally, in Expert Consultation, the knowledge from $T$ experts is aggregated using the $T$ weight vectors and is added to the common knowledge, producing the final projection vectors to embed the bag of patches $\mathbf{x}$ in the model space.
Given an input WSI from a target $t^{th}$ task, this process can be understood as a consultation with $T$ experts, where the $t^{th}$ expert has the best knowledge of the task and thus contributes the most to the final decision.

\subsubsection{Preliminary Consultation.} The router $\mathcal{R}(\cdot)$ is a stack of two linear layers. A \texttt{ReLU} activation is placed between the two layers. 
Given an input sequence of patch embeddings $\mathbf{x}$, the first layer maps the dimension size from $d_f$ to $d_{model}$ and the second layer produces a set of $T$ weight vectors $\mathcal{W} = \{\mathbf{w}_i\}_{i=1}^{T} = FC_2\big(\texttt{ReLU}(FC_1(\mathbf{x})\big)$, where $FC_1$ and $FC_2$ denote two linear layers and $\mathbf{w}_i \in \mathbb{R}^{N}$ is a weight vector for all $N$ patches in a WSI. 



\subsubsection{Expert Assignment.} We represent the knowledge of each expert $\tau_i$ as a learnable matrix $\mathbb{R}^{d_{f}\times d_{model}}$. 
Along with the input WSI, the task indicator $\mathcal{T}$ specifies its task by assigning 1 to the target task and 0 to others. Using $\mathcal{T}$, we increase the weights of the target task $\mathbf{w}_t$ and decrease the weights of other tasks by scaling and shifting procedures. The scaling procedure is given by:


\begin{equation}
    \tilde{\mathcal{W}} = \{\tilde{\mathbf{w}}_i\}^{T}_{i=1}, \text{ where } \tilde{\mathbf{w}}_i = \frac{\exp\big(\mathbbm{1}_{i=t}(\mathbf{w}_i \cdot \mathbf{1}\gamma) + \mathbbm{1}_{i\neq t}(\mathbf{w}_i)\big)}{\sum^{T}_{i \neq t} \exp(\mathbf{w}_i) + \mathbf{w}_t \cdot \mathbf{1}\gamma }
    \label{eq:3}
\end{equation}

\noindent where $\gamma$ is a scaling hyper-parameter.
In the shifting procedure, we first take the mean value of $\mathbf{\tilde{w}}_i$ to obtain a scalar weight for each task. Then, we shift the target weight to further distinguish it from other tasks. The procedure can be formulated as follows:

\begin{equation}
\bar{\mathbf{w}} = \{\bar{{w}}_i\}^{T}_{i=1}, \text{ where } \bar{w}_i = \frac{1}{N} \sum^{N}_{j=1} \tilde{w}_{ij} + \mathbbm{1}_{i=t} \cdot \beta,
\label{eq:4}
\end{equation}

\noindent where $\beta$ is a shifting hyper-parameter and $\tilde{w}_{ij}$ is the scalar weight of the $i^{th}$ task for the $j^{th}$ patch in a bag of patches. In this manner, the expert $\tau_t$ is assigned the larger weight and others have smaller weights, which are determined by $\mathcal{R}(\cdot)$.

\subsubsection{Expert Consultation.}
We combine the knowledge of all experts $\boldsymbol{\tau}$ using $\bar{\mathbf{w}}$, i.e., computing a weighted sum. Then, the combined knowledge is added to the common knowledge of all tasks $\theta_p \in \mathbb{R}^{d_f \times d_{model}}$, which is also a learnable matrix, producing the final projection vectors $\theta^{*}_p$ as follows:


\begin{equation}
\begin{array}{ll}
\theta^{*}_p = \theta_p + \sum^{T}_{i=1} \tau_i \cdot \bar{w}_i.
\end{array}
\end{equation}

\noindent $\theta^{*}_p$ is used to project $\mathbf{x}$ as $\mathbf{v^{(0)}} = \mathcal{P}_{\texttt{ECN}}(\mathbf{x}, \theta^{*}_p)$, aiming to produce the most suitable representation by encompassing the specialized knowledge from the target task and other tasks and the shared knowledge among all tasks.
Then, $\mathbf{v^{(0)}}$ is processed by the encoder $\mathcal{E}(\cdot)$ and the decoder $\mathcal{D}(\cdot)$.


\subsection{Visual Encoder} 

The encoder $\mathcal{E}(\cdot)$ consists of $L_e$ encoder layers. Each layer includes a normalization layer, denoted as $\text{LN}(\cdot)$, followed by the Nystr\"om attention mechanism $\text{NA}(\cdot)$ \cite{xiong2021Nystromformer}. $\text{NA}(\cdot)$ approximates self-attention operation for computational efficiency since it needs to process thousands of tokens from a WSI. 
A skip connection adds the input of $\text{LN}(\cdot)$ to the output of $\text{NA}(\cdot)$. This process for each $l^{th}$ encoder layer can be formulated as follows:

\begin{equation}
\mathbf{v}^{(l)} = \mathbf{v}^{(l-1)} + \text{NA}\big(\text{LN}(\mathbf{v}^{(l-1)})\big), 1 \leq l \leq L_e,
\label{eq:encoder}
\end{equation}

\noindent where $\mathbf{v}^{(0)} = \mathcal{P}_{\texttt{ECN}}(\mathbf{x}, \theta^{*}_p)$.

\subsection{Language Decoder} 

There are $ L_d $ decoder layers in the decoder $\mathcal{D}(\cdot)$. Each decoder layer receives two inputs: one is the visual token embeddings from $\mathcal{E}(\cdot)$ and the other is the word embeddings from the previous word generated by $\mathcal{D}(\cdot)$. To produce the word embeddings, we adopt a word embedding layer $\text{WE}(\cdot,\theta_w)$ that is used to embed a set of natural words $\mathbf{s} = \{s_i\}_{i=1}^{S}$ (represented as token IDs) into the model space. $\theta_w \in \mathbb{R}^{N_{\text{voc}} \times d_{\text{model}}}$ and $N_{\text{voc}}$ is the number words in the vocabulary. Then, a positional encoding $\text{PE}$ \cite{vaswani2017attention} is added given by $\mathbf{h}^{(0)} = \text{WE}(\mathbf{s}) + \text{PE}$.

Each $l^{th}$ decoder layer contains a series of masked multi-head self-attention ($\text{MHSA}$) and multi-head cross-attention ($\text{MHCA}$), of which each is preceded by $\text{LN}(\cdot)$. 
Masked $\text{MHSA}$ learns the intra-relevance of the hidden tokens $\mathbf{h}^{(l)}$ without utilizing the future ground truth words during training.
$\text{MHCA}$ aligns the hidden tokens and visual token embeddings produced by $\mathcal{E}(\cdot)$. 
$\text{MHCA}$ is followed by a point-wise feed-forward network ($\text{PWFF}$).
We note that the number of tokens in $\mathbf{h}^{(l)}$ is much smaller than the number of tokens (patches) in $\mathcal{E}(\cdot)$. Hence, vanilla $\text{MHSA}$ \cite{vaswani2017attention} is used for both. 
The process of the $l^{th}$ decoder layer can be formulated as follows:

\begin{equation}
\begin{array}{ll}
\mathbf{h}^{(l)} = \mathbf{h}^{(l-1)} + \text{Masked\_MHSA}\big(\text{LN}(\mathbf{h}^{(l-1)})\big), \\
\mathbf{h}^{(l)} = \mathbf{h}^{(l)} + \text{MHCA}\big(\text{LN}(\mathbf{h}^{(l)}), \mathbf{v}\big), \\
\mathbf{h}^{(l)} = \text{PWFF}(\mathbf{h}^{(l)})
\end{array}, 1 \leq l \leq L_d.
\end{equation}



Following the decoder $\mathcal{D}(\cdot)$, a linear classifier $\text{FC}(\cdot, \theta_{c})$ where $\theta_{c} \in \mathbb{R}^{d_{model}\times N_{voc}}$ is designed to produce logits $\hat{\mathbf{p}} = \text{FC}(\mathbf{h})$ where $\hat{\mathbf{p}} \in \mathbb{R}^{N_{voc}}$.

\section{Experimental Results}

\subsection{Dataset}
\begin{wrapfigure}{r}{0.5\textwidth}
\centerline{\includegraphics[width=6cm]{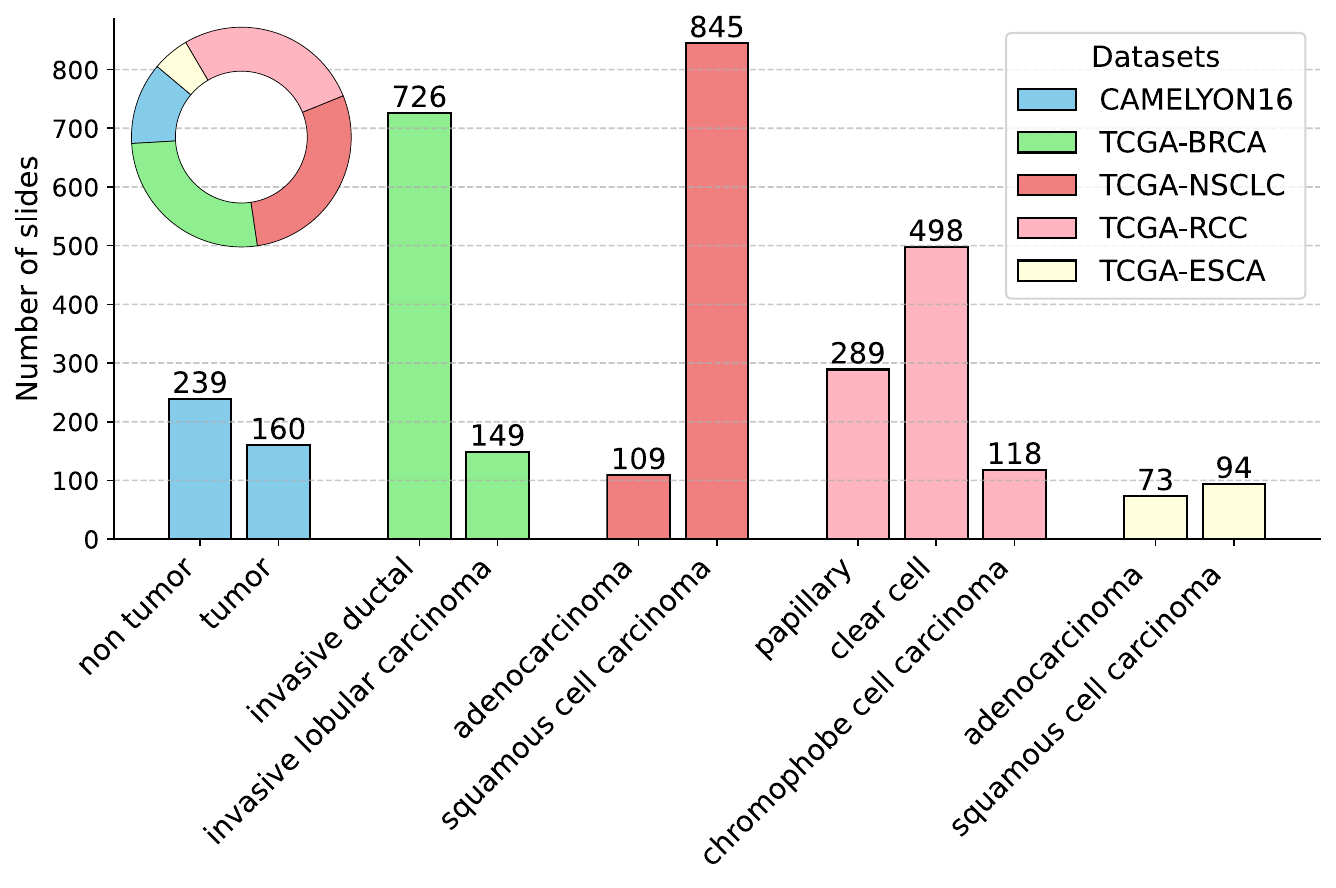}}
\caption{Distribution of five datasets: CAMELYON16, TCGA-BRCA, TCGA-ESCA, TCGA-RCC and TCGA-NSCLC.}
\label{fig:dataset}
\end{wrapfigure}

In this study, we consider five datasets/tasks: a breast tumor classification task (CAMELYON16) and four cancer subtyping tasks from four organs, including breast (TCGA-BRCA), esophagus (TCGA-ESCA), lung (TCGA-NSCLC), and kidney (TCGA-RCC). Figure \ref{fig:dataset} shows the distribution of the WSIs in the five datasets. To ensure the validity of our experiments, we conduct each task three times with three different train-validation-test splits under the same random seed. For each run, the best checkpoint on the validation set is used to evaluate the test dataset.

\subsection{Implemental Details} The embedding size $ d_{model} $ is set to 512. The number of both encoder and decoder layers ($ L_e $ and $ L_d $) is set to 2. For the generative decoding process, we build a vocabulary including 18 words, including $\texttt{<BOS>}$ and $\texttt{<EOS>}$ tokens; hence, $ N_{voc} = 18 $. For Nystr\"om Attention and Multi-head Self-attention, we set the number of heads $ h = 8 $. Both scaling and shifting hyperparameters ($\gamma$ and $\beta$) are set to 5. We train MECFormer for 200 epochs, using early stopping when the validation loss does not decrease for five consecutive epochs. \texttt{Lookahead RAdam} \cite{zhang2019lookahead} is used as the optimizer, with the learning rate set to $ 1e^{-5} $ and cross-entropy loss function. All the experiments are conducted on an NVIDIA A6000 GPU.

\subsection{Comparative Study} 
\subsubsection{Competitors.} We compare MECFormer to four other MIL-based models for WSI analysis: CLAM-MB \cite{lu2021data}, TransMIL \cite{shao2021transmil}, DSMIL \cite{dsmil}, and DGCN \cite{dgcn}. CLAM-MB utilizes an attention mechanism and an auxiliary task on cropped patches to enhance feature representation. TransMIL proposes a learnable convolution network for positional encoding between transformer layers. 
DSMIL uses an instance classifier to identify the critical instance, a non-local attention mechanism to aggregate all instances along with the critical instance, and a bag classifier for the prediction. 
DGCN constructs a graph for a WSI based on distances between pairs of patches and adopts a graph convolution network for the classification. 

\subsubsection{Comparison setting.} 
We compare MECFormer with four models under two settings: 1) \textit{individual training}, where each model is trained separately for each task, and 2) \textit{joint training}, where all five tasks are combined into a single multi-class classification. For MECFormer, we use \textit{joint training} $+\mathcal{T}$, meaning the model is aware of the target task indicated by $\mathcal{T}$. Both MECFormer and the other models use two pre-trained feature extractors: CTransPath \cite{wang2022transformer} and UNI \cite{chen2024towards}.


\subsubsection{Evaluation metrics.} For evaluation, we use four metrics: Accuracy (Acc), F1 score (F1), Recall, and Precision. In the \textit{joint training} setting, models may predict \textit{invalid terms}, i.e., terms not in the ground-truth categories for a task. The overall metric is computed as $\frac{\sum^{N_{c}}_{i=1} {m_i}}{N_{c} + N_{o}}$, where $m_i$ is F1, Precision, or Recall for each category, $N_{c}$ is the number of ground-truth categories, and $N_{o}$ is the number of out-of-distribution categories, further penalizing misclassifications.

\subsection{Main Results} 
We report the classification results of MECFormer and the comparison to the state-of-the-art MIL models in Table \ref{tab:main_ctrans} and \ref{tab:main_uni}. 

\subsubsection{Results on CTransPath.} As shown in Table \ref{tab:main_ctrans}, using CTransPath as $\mathcal{G}(\cdot)$, MECFormer was shown to be the best performer over the four competing models under \textit{individual training}. 
For CAMELYON16, TCGA-ESCA, and TCGA-NSCLC, MECFormer achieves the best performance regardless of the evaluation metrics with the improvement of $\geq$1.291\% Acc, $\geq$0.013 F1, $\geq$0.007 Recall, and $\geq$0.023 Precision for CAMELYON16, 
$\geq$0.412\% Acc, $\geq$ 0.043 F1, $\geq$0.037 Recall, and $\geq$0.036 Precision for TCGA-ESCA, and 
$\geq$0.412\% Acc and $\geq$0.004 F1, $\geq$0.003 Recall, and $\geq$0.003 Precision for TCGA-NSCLC. 
As for the remaining tasks, MECFormer was superior to other competitors except for Recall in TCGA-BRCA and Recall and Precision in TCGA-RCC. For Recall and Precision in TCGA-RCC, MECFormer was the second-best model, while it was inferior to DSMIL in Recall in TCGA-BRCA, which obtained the second-best Recall of 0.904. It is noteworthy that none of the competitors were able to outperform MECFormer in both Recall and Precision, which explains MECFormer's superior F1.
In \textit{joint training}, the four competitors obtained lower performance than their own performance in \textit{individual training}. This is due to the invalid (or out-of-distribution) predictions made by the four competitors. These results suggest that naive multi-task learning is not specific enough to handle multiple tasks simultaneously. 
MECFormer, however, was not sensitive to the presence of out-of-distribution categories, resulting in a substantial performance gap; for instance, MECFormer improved F1 by 0.121$\sim$0.413, 0.121$\sim$0.329, 0.179$\sim$0.408, and 0.311$\sim$0.696 in comparison to CLAM, TransMIL, DSMIL, and DGCN, respectively, across the five tasks.

\subsubsection{Results on UNI.} 
Employing UNI as $\mathcal{G}(\cdot)$, we made similar observations with the above experiments with CTransPath. MECFormer demonstrated superior performance compared to the four competing models under \textit{individual training} with the exception of TCGA-ESCA. 
In TCGA-ESCA, MECFormer achieved the second-best results for Acc, F1, and Recall, while CLAM obtained the best performance across all metrics. 
For CAMELYON16, TCGA-NSCLC, and TCGA-RCC, MECFormer was superior to other competitors regardless of the evaluation metrics; for instance, $\geq$0.517\% Acc, $\geq$0.006 F1, $\geq$0.001 Recall, and $\geq$0.010 Precision for CAMELYON16, $\geq$0.823\% Acc, $\geq$0.008 F1, $\geq$0.008 Recall, and $\geq$0.008 Precision for TCGA-NSCLC, and $\geq$0.383\% Acc, $\geq$0.001 F1, $\geq$0.004 Recall, and $\geq$0.003 in Precision for TCGA-RCC. 
As for TCGA-BRCA, MECFormer improved upon other four models on Acc, F1, and Precision by $\geq$0.390\%, $\geq$0.002, and $\geq$0.018, respectively. With respect to Recall, it was inferior to CLAM, which obtained a Recall of 0.912.
Under \textit{joint training}, numerous invalid predictions were made by the four other models similar to the results with CTransPath. For this reason, the four competitors were substantially inferior to MECFormer; for example, F1 was dropped by 0.266$\sim$0.596, 0.218$\sim$0.345, 0.245$\sim$0.649, and 0.280$\sim$0.661 for CLAM, TransMIL, DSMIL, and DGCN, respectively, in comparison to MECFormer.

\begin{table}[]
    \centering
    \caption{Classification results on five tasks using CTransPath as a feature extractor. The best and second-best values are highlighted in \textcolor{red}{red} and \textcolor{blue}{blue}, respectively.}
    \resizebox{1.0\textwidth}{!}{%
    \begin{tabular}{P{3cm}|P{3cm}|P{3cm}|P{3cm}|P{3cm}|P{3cm}}
    \toprule
    \textbf{Setting} & \textbf{Method} & \textbf{Acc (\%)}    & \textbf{F1}   & \textbf{Recall} & \textbf{Precision} \\ \midrule
    \multicolumn{6}{c}{\textbf{CAMELYON16}}                                                           \\ \midrule
    \multirow{4}{*}{Individual training}  & CLAM-MB \cite{lu2021data}              & 91.731 ($\pm$3.133)  & 0.911 ($\pm$0.034) & 0.906 ($\pm$0.033)   & 0.918 ($\pm$0.034)      \\
    & TransMIL \cite{shao2021transmil}       & \textcolor{blue}{93.024 ($\pm$1.343)}  & \textcolor{blue}{0.925 ($\pm$0.014)} & \textcolor{blue}{0.919 ($\pm$0.013)}   & \textcolor{blue}{0.933 ($\pm$0.016)}      \\
    & DSMIL \cite{dsmil}                     & 90.439 ($\pm$0.448)  & 0.897 ($\pm$0.005) & 0.893 ($\pm$0.005)   & 0.903 ($\pm$0.005)      \\
    & DGCN \cite{dgcn}                       & 90.698 ($\pm$3.379)  & 0.897 ($\pm$0.039) & 0.884 ($\pm$0.040)   & 0.921 ($\pm$0.032)      \\ \midrule
    \multirow{4}{*}{Joint training} 
    & CLAM-MB \cite{lu2021data}              & 92.765 ($\pm$1.184)  & 0.817 ($\pm$0.168) & 0.805 ($\pm$0.165) & 0.839 ($\pm$0.176)      \\
    & TransMIL \cite{shao2021transmil}       & 89.922 ($\pm$0.775)  & 0.793 ($\pm$0.170) & 0.782 ($\pm$0.160) & 0.819 ($\pm$0.189)      \\
    & DSMIL \cite{dsmil}                     & 84.496 ($\pm$10.769)  & 0.759 ($\pm$0.241) & 0.755 ($\pm$0.225) & 0.777 ($\pm$0.247)      \\
    & DGCN \cite{dgcn}                       & 62.016 ($\pm$0.000)  & 0.242 ($\pm$0.026) & 0.307 ($\pm$0.046) & 0.234 ($\pm$0.044)      \\ \midrule
    Joint Training + $\mathcal{T}$ 
    & MECFormer (ours)                                   & \textcolor{red}{94.315 ($\pm$1.614)}  & \textcolor{red}{0.938 ($\pm$0.018)} & \textcolor{red}{0.926 ($\pm$0.020)}   & \textcolor{red}{0.956 ($\pm$0.014)} \\ \midrule
    \multicolumn{6}{c}{\textbf{TCGA--BRCA}} \\ \midrule
    \multirow{4}{*}{Individual training} & CLAM-MB \cite{lu2021data}              & \textcolor{blue}{94.103 ($\pm$1.642)}  & \textcolor{blue}{0.896 ($\pm$0.030)} & 0.896 ($\pm$0.067)   & \textcolor{blue}{0.911 ($\pm$0.053)} \\
    & TransMIL \cite{shao2021transmil}       & 93.663 ($\pm$2.674)  & 0.887 ($\pm$0.043) & \textcolor{red}{0.914 ($\pm$0.070)}   & 0.871 ($\pm$0.039)      \\
    & DSMIL \cite{dsmil}                     & 92.813 ($\pm$2.834)  & 0.884 ($\pm$0.031) & \textcolor{blue}{0.904 ($\pm$0.035)}   & 0.878 ($\pm$0.058)      \\
    & DGCN \cite{dgcn}                       & 92.027 ($\pm$1.820)  & 0.865 ($\pm$0.026) & 0.873 ($\pm$0.054)   & 0.861 ($\pm$0.007)      \\ \midrule
    \multirow{4}{*}{Joint training} 
    & CLAM-MB \cite{lu2021data}              & 92.006 ($\pm$1.527)  & 0.543 ($\pm$0.085) & 0.551 ($\pm$0.104) & 0.539 ($\pm$0.073)      \\
    & TransMIL \cite{shao2021transmil}       & 91.957 ($\pm$5.269)  & 0.635 ($\pm$0.277) & 0.630 ($\pm$0.308) & 0.660 ($\pm$0.239)      \\
    & DSMIL \cite{dsmil}                     & 91.963 ($\pm$3.162)  & 0.538 ($\pm$0.076) & 0.527 ($\pm$0.072) & 0.563 ($\pm$0.088)      \\
    & DGCN \cite{dgcn}                       & 88.152 ($\pm$6.000)  & 0.512 ($\pm$0.048) & 0.510 ($\pm$0.056) & 0.524 ($\pm$0.056)      \\ \midrule
    Joint Training + $\mathcal{T}$ 
    & MECFormer (ours)                                   & \textcolor{red}{94.512 ($\pm$2.712)}  & \textcolor{red}{0.903 ($\pm$0.047)} & 0.898 ($\pm$0.077) & \textcolor{red}{0.916 ($\pm$0.026)}      \\ \midrule
    \multicolumn{6}{c}{\textbf{TCGA--ESCA}} \\ \midrule
    \multirow{4}{*}{Individual training} & CLAM-MB \cite{lu2021data}              & 89.794 ($\pm$5.836)  & 0.895 ($\pm$0.061) & \textcolor{blue}{0.903 ($\pm$0.061)}   & 0.894 ($\pm$0.059)      \\
    & TransMIL \cite{shao2021transmil}       & \textcolor{blue}{89.805 ($\pm$2.193)}  & 0.895 ($\pm$0.022) & 0.892 ($\pm$0.020) & \textcolor{blue}{0.905 ($\pm$0.030)}      \\
    & DSMIL \cite{dsmil}                     & 89.805 ($\pm$5.789)  & 0.895 ($\pm$0.061) & 0.894 ($\pm$0.066)   & 0.904 ($\pm$0.050)      \\
    & DGCN \cite{dgcn}                       & 89.805 ($\pm$5.789)  & \textcolor{blue}{0.896 ($\pm$0.060)} & 0.896 ($\pm$0.064)   & 0.899 ($\pm$0.054)      \\ \midrule
    \multirow{4}{*}{Joint training} 
    & CLAM-MB \cite{lu2021data}              & 86.180 ($\pm$9.329)  & 0.530 ($\pm$0.113) & 0.523 ($\pm$0.117) & 0.545 ($\pm$0.105)      \\
    & TransMIL \cite{shao2021transmil}       & 91.602 ($\pm$5.782)  & 0.818 ($\pm$0.198) & 0.816 ($\pm$0.202) & 0.819 ($\pm$0.195)      \\
    & DSMIL \cite{dsmil}                     & 85.628 ($\pm$3.575)  & 0.531 ($\pm$0.102) & 0.521 ($\pm$0.102) & 0.546 ($\pm$0.094)      \\
    & DGCN \cite{dgcn}                       & 78.994 ($\pm$9.385)  & 0.487 ($\pm$0.113) & 0.472 ($\pm$0.110) & 0.532 ($\pm$0.099)      \\ \midrule
    Joint Training + $\mathcal{T}$
    & MECFormer (ours)                                   & \textcolor{red}{94.004 ($\pm$3.739)}  & \textcolor{red}{0.939 ($\pm$0.038)} & \textcolor{red}{0.940 ($\pm$0.041)}   & \textcolor{red}{0.941 ($\pm$0.034)}      \\ \midrule
    \multicolumn{6}{c}{\textbf{TCGA--NSCLC}} \\ \midrule
    \multirow{4}{*}{Individual training} & CLAM-MB \cite{lu2021data}              & 90.606 ($\pm$1.843)  & 0.906 ($\pm$0.018) & 0.907 ($\pm$0.015)   & 0.911 ($\pm$0.019)      \\
    & TransMIL \cite{shao2021transmil}       & \textcolor{blue}{92.550 ($\pm$2.874)}  & \textcolor{blue}{0.925 ($\pm$0.029)} & \textcolor{blue}{0.927 ($\pm$0.027)} & \textcolor{blue}{0.929 ($\pm$0.027)} \\
    & DSMIL \cite{dsmil}                     & 89.017 ($\pm$2.876)  & 0.890 ($\pm$0.029) & 0.893 ($\pm$0.026)   & 0.895 ($\pm$0.024)      \\
    & DGCN \cite{dgcn}                       & 85.128 ($\pm$2.055)  & 0.851 ($\pm$0.020) & 0.852 ($\pm$0.019)   & 0.855 ($\pm$0.021)      \\ \midrule
    \multirow{4}{*}{Joint training} 
    & CLAM-MB \cite{lu2021data}              & 87.569 ($\pm$4.920)  & 0.516 ($\pm$0.155) & 0.511 ($\pm$0.155) & 0.525 ($\pm$0.149)      \\
    & TransMIL \cite{shao2021transmil}       & 89.073 ($\pm$2.060)  & 0.600 ($\pm$0.014) & 0.595 ($\pm$0.013) & 0.606 ($\pm$0.014)      \\
    & DSMIL \cite{dsmil}                     & 90.194 ($\pm$0.578)  & 0.559 ($\pm$0.090) & 0.551 ($\pm$0.088) & 0.569 ($\pm$0.093)      \\
    & DGCN \cite{dgcn}                       & 84.745 ($\pm$1.850)  & 0.526 ($\pm$0.075) & 0.518 ($\pm$0.083) & 0.540 ($\pm$0.070)      \\ \midrule
    Joint Training + $\mathcal{T}$ 
    & MECFormer (ours)                                   & \textcolor{red}{92.962 ($\pm$1.880)}  & \textcolor{red}{0.929 ($\pm$0.019)} & \textcolor{red}{0.930 ($\pm$0.017)}   & \textcolor{red}{0.932 ($\pm$0.020)}      \\ \midrule
    \multicolumn{6}{c}{\textbf{TCGA--RCC}}   \\ \midrule
    \multirow{4}{*}{Individual training} & CLAM-MB \cite{lu2021data}              & 95.788 ($\pm$1.330)  & 0.945 ($\pm$0.021) & 0.935 ($\pm$0.011)   & \textcolor{red}{0.958 ($\pm$0.039)}      \\
    & TransMIL \cite{shao2021transmil}       & 95.393 ($\pm$3.031)  & 0.942 ($\pm$0.033) & 0.936 ($\pm$0.042)   & 0.955 ($\pm$0.028)      \\
    & DSMIL \cite{dsmil}                     & \textcolor{red}{96.160 ($\pm$1.312)}  & \textcolor{blue}{0.950 ($\pm$0.012)} & \textcolor{red}{0.966 ($\pm$0.006)}   & 0.937 ($\pm$0.022)      \\
    & DGCN \cite{dgcn}                       & 93.861 ($\pm$3.303)  & 0.931 ($\pm$0.043) & 0.949 ($\pm$0.023)   & 0.921 ($\pm$0.058)      \\ \midrule
    \multirow{4}{*}{Joint training} 
    & CLAM-MB \cite{lu2021data}              & 93.478 ($\pm$4.023)  & 0.772 ($\pm$0.134) & 0.785 ($\pm$0.155) & 0.765 ($\pm$0.120)      \\
    & TransMIL \cite{shao2021transmil}       & 91.936 ($\pm$3.010)  & 0.602 ($\pm$0.065) & 0.603 ($\pm$0.085) & 0.605 ($\pm$0.049)      \\
    & DSMIL \cite{dsmil}                     & 91.932 ($\pm$1.952)  & 0.645 ($\pm$0.082) & 0.621 ($\pm$0.083) & 0.621 ($\pm$0.059)      \\
    & DGCN \cite{dgcn}                       & 92.315 ($\pm$2.366)  & 0.646 ($\pm$0.078) & 0.646 ($\pm$0.080) & 0.647 ($\pm$0.080)      \\ \midrule
    Joint Training + $\mathcal{T}$ & MECFormer (ours)                                   & \textcolor{red}{96.160 ($\pm$1.744)}  & \textcolor{red}{0.957 ($\pm$0.032)} & \textcolor{blue}{0.960 ($\pm$0.030)}   & \textcolor{blue}{0.957 ($\pm$0.035)}  \\
    \bottomrule
    \end{tabular}}
    \label{tab:main_ctrans}
    \end{table}

\begin{table}[]
    \centering
    \caption{Classification results on five tasks using UNI as a feature extractor. The best and second-best values are highlighted in \textcolor{red}{red} and \textcolor{blue}{blue}, respectively.}
    \resizebox{1.0\textwidth}{!}{%
    \begin{tabular}{P{3cm}|P{3cm}|P{3cm}|P{3cm}|P{3cm}|P{3cm}}
    \toprule
    \textbf{Setting} & \textbf{Method} & \textbf{Acc (\%)}    & \textbf{F1}   & \textbf{Recall} & \textbf{Precision} \\ \midrule
    \multicolumn{6}{c}{\textbf{CAMELYON16}}                                                           \\ \midrule
    \multirow{4}{*}{Individual training} & CLAM-MB \cite{lu2021data}        & \textcolor{blue}{97.674 ($\pm$1.550)}  & \textcolor{blue}{0.975 ($\pm$0.017)} & \textcolor{blue}{0.975 ($\pm$0.018)} & \textcolor{blue}{0.976 ($\pm$0.015)}      \\
    & TransMIL \cite{shao2021transmil}       & 92.506 ($\pm$1.614)  & 0.917 ($\pm$0.019) & 0.901 ($\pm$0.021)   & 0.946 ($\pm$0.010)      \\
    & DSMIL \cite{dsmil}          & 93.540 ($\pm$3.663)  & 0.929 ($\pm$0.041) & 0.918 ($\pm$0.045)   & 0.948 ($\pm$0.032)      \\
    & DGCN \cite{dgcn}           & 78.295 ($\pm$10.137) & 0.739 ($\pm$0.130) & 0.734 ($\pm$0.119)   & 0.823 ($\pm$0.115)      \\ \midrule
    \multirow{4}{*}{Joint training} 
    & CLAM-MB \cite{lu2021data}              & 92.506 ($\pm$4.408)  & 0.385 ($\pm$0.089) & 0.374 ($\pm$0.092) & 0.400 ($\pm$0.081)      \\
    & TransMIL \cite{shao2021transmil}       & 95.349 ($\pm$0.775)  & 0.636 ($\pm$0.005) & 0.627 ($\pm$0.007) & 0.647 ($\pm$0.004)      \\
    & DSMIL \cite{dsmil}                     & 73.902 ($\pm$2.087)  & 0.332 ($\pm$0.192) & 0.327 ($\pm$0.257) & 0.353 ($\pm$0.246)      \\
    & DGCN \cite{dgcn}                       & 62.016 ($\pm$0.000)  & 0.320 ($\pm$0.109) & 0.417 ($\pm$0.144) & 0.259 ($\pm$0.088)      \\ \midrule
    Joint Training + $\mathcal{T}$ 
    & MECFormer (ours)   & \textcolor{red}{98.191 ($\pm$1.630)}  & \textcolor{red}{0.981 ($\pm$0.024)} & \textcolor{red}{0.976 ($\pm$0.037)}   & \textcolor{red}{0.986 ($\pm$0.055)} \\ \midrule
    \multicolumn{6}{c}{\textbf{TCGA--BRCA}}                                                            \\ \midrule
    \multirow{4}{*}{Individual training} & CLAM-MB \cite{lu2021data}        & \textcolor{blue}{93.700 ($\pm$1.215)}  & \textcolor{blue}{0.895 ($\pm$0.018)} & \textcolor{red}{0.912 ($\pm$0.056)}   & 0.886 ($\pm$0.012)      \\
    & TransMIL \cite{shao2021transmil}       & 93.262 ($\pm$0.913)  & 0.884 ($\pm$0.019) & 0.891 ($\pm$0.064)   & \textcolor{blue}{0.890 ($\pm$0.040)}      \\
    & DSMIL \cite{dsmil}          & 92.439 ($\pm$1.213)  & 0.872 ($\pm$0.018) & 0.885 ($\pm$0.051)   & 0.868 ($\pm$0.016)      \\
    & DGCN \cite{dgcn}            & 91.589 ($\pm$0.820)  & 0.860 ($\pm$0.016) & 0.883 ($\pm$0.076)   & 0.860 ($\pm$0.044)      \\ \midrule
    \multirow{4}{*}{Joint training} 
    & CLAM-MB \cite{lu2021data}              & 89.441 ($\pm$3.227)  & 0.480 ($\pm$0.081) & 0.475 ($\pm$0.061) & 0.499 ($\pm$0.120)      \\
    & TransMIL \cite{shao2021transmil}       & 91.584 ($\pm$0.890)  & 0.677 ($\pm$0.159) & 0.699 ($\pm$0.216) & 0.677 ($\pm$0.125)      \\
    & DSMIL \cite{dsmil}                     & 82.387 ($\pm$5.500)  & 0.541 ($\pm$0.077) & 0.585 ($\pm$0.074) & 0.583 ($\pm$0.064)      \\
    & DGCN \cite{dgcn}                       & 88.623 ($\pm$3.495)  & 0.450 ($\pm$0.075) & 0.428 ($\pm$0.054) & 0.520 ($\pm$0.136)      \\ \midrule
    Joint Training + $\mathcal{T}$ 
    & MECFormer (ours)   & \textcolor{red}{94.090 ($\pm$0.448)}  & \textcolor{red}{0.897 ($\pm$0.005)} & \textcolor{blue}{0.893 ($\pm$0.006)}   & \textcolor{red}{0.908 ($\pm$0.003)}      \\ \midrule
    \multicolumn{6}{c}{\textbf{TCGA--ESCA}}                                                            \\ \midrule
    \multirow{4}{*}{Individual training} & CLAM-MB \cite{lu2021data}        & \textcolor{red}{95.801 ($\pm$1.087)} & \textcolor{red}{0.958 ($\pm$0.011)} & \textcolor{red}{0.960 ($\pm$0.007)} & \textcolor{red}{0.957 ($\pm$0.011)}      \\
    & TransMIL \cite{shao2021transmil}       & 91.017 ($\pm$3.573)  & 0.908 ($\pm$0.038) & 0.906 ($\pm$0.040)   & 0.915 ($\pm$0.029)      \\
    & DSMIL \cite{dsmil}          & 92.792 ($\pm$3.669)  & 0.927 ($\pm$0.036) & 0.929 ($\pm$0.031)   & \textcolor{blue}{0.935 ($\pm$0.032)}      \\
    & DGCN \cite{dgcn}            & 87.446 ($\pm$3.491)  & 0.871 ($\pm$0.037) & 0.870 ($\pm$0.042)   & 0.878 ($\pm$0.032)      \\ \midrule
    \multirow{4}{*}{Joint training} 
    & CLAM-MB \cite{lu2021data}              & 90.985 ($\pm$4.872)  & 0.510 ($\pm$0.053) & 0.501 ($\pm$0.056) & 0.520 ($\pm$0.051)      \\
    & TransMIL \cite{shao2021transmil}       & 86.829 ($\pm$4.481)  & 0.669 ($\pm$0.122) & 0.666 ($\pm$0.121) & 0.688 ($\pm$0.145)      \\
    & DSMIL \cite{dsmil}                     & 86.212 ($\pm$5.803)  & 0.466 ($\pm$0.146) & 0.453 ($\pm$0.156) & 0.486 ($\pm$0.131)      \\
    & DGCN \cite{dgcn}                       & 80.271 ($\pm$9.205)  & 0.438 ($\pm$0.004) & 0.427 ($\pm$0.006) & 0.493 ($\pm$0.070)      \\ \midrule
    Joint Training + $\mathcal{T}$ 
    & MECFormer (ours)   & \textcolor{blue}{93.398 ($\pm$2.799)}  & \textcolor{blue}{0.934 ($\pm$0.028)} & \textcolor{blue}{0.939 ($\pm$0.026)} & 0.934 ($\pm$0.025)      \\ \midrule
    \multicolumn{6}{c}{\textbf{TCGA--NSCLC}}                                                           \\ \midrule
    \multirow{4}{*}{Individual training} & CLAM-MB \cite{lu2021data}        & 91.457 ($\pm$3.877)  & 0.914 ($\pm$0.039) & 0.916 ($\pm$0.037)   & 0.919 ($\pm$0.035)      \\
    & TransMIL \cite{shao2021transmil}       & \textcolor{blue}{92.962 ($\pm$2.970)}  & \textcolor{blue}{0.930 ($\pm$0.030)} & \textcolor{blue}{0.931 ($\pm$0.027)}   & \textcolor{blue}{0.932 ($\pm$0.026)}      \\
    & DSMIL \cite{dsmil}          & 91.457 ($\pm$5.557)  & 0.914 ($\pm$0.055) & 0.915 ($\pm$0.054)   & 0.918 ($\pm$0.056)      \\
    & DGCN \cite{dgcn}            & 91.315 ($\pm$2.179)  & 0.913 ($\pm$0.022) & 0.913 ($\pm$0.022)   & 0.915 ($\pm$0.023)      \\ \midrule
    \multirow{4}{*}{Joint training} 
    & CLAM-MB \cite{lu2021data}              & 91.401 ($\pm$4.712)  & 0.672 ($\pm$0.262) & 0.668 ($\pm$0.264) & 0.677 ($\pm$0.259)      \\
    & TransMIL \cite{shao2021transmil}       & 90.634 ($\pm$2.049)  & 0.706 ($\pm$0.155) & 0.702 ($\pm$0.157) & 0.710 ($\pm$0.152)      \\
    & DSMIL \cite{dsmil}                     & 84.249 ($\pm$2.952)  & 0.531 ($\pm$0.304) & 0.521 ($\pm$0.313) & 0.550 ($\pm$0.304)      \\
    & DGCN \cite{dgcn}                       & 83.624 ($\pm$3.490)  & 0.427 ($\pm$0.015) & 0.420 ($\pm$0.017) & 0.441 ($\pm$0.009)      \\ \midrule
    Joint Training + $\mathcal{T}$ 
    & MECFormer (ours)   & \textcolor{red}{93.785 ($\pm$2.776)}  & \textcolor{red}{0.938 ($\pm$0.028)} & \textcolor{red}{0.939 ($\pm$0.026)}   & \textcolor{red}{0.940 ($\pm$0.025)}      \\ \midrule
    \multicolumn{6}{c}{\textbf{TCGA--RCC}}                                                             \\ \midrule
    \multirow{4}{*}{Individual training} & CLAM-MB \cite{lu2021data}        & 95.015 ($\pm$3.310)  & 0.938 ($\pm$0.045) & 0.955 ($\pm$0.034)   & 0.927 ($\pm$0.058)      \\
    & TransMIL \cite{shao2021transmil}       & \textcolor{blue}{96.547 ($\pm$2.292)}  & \textcolor{blue}{0.958 ($\pm$0.035)} & \textcolor{blue}{0.966 ($\pm$0.030)}   & \textcolor{blue}{0.948 ($\pm$0.045)}      \\
    & DSMIL \cite{dsmil}          & 95.384 ($\pm$1.150)  & 0.941 ($\pm$0.020) & 0.942 ($\pm$0.015)   & 0.941 ($\pm$0.026)      \\
    & DGCN \cite{dgcn}            & 93.086 ($\pm$2.272)  & 0.912 ($\pm$0.024) & 0.920 ($\pm$0.014)   & 0.913 ($\pm$0.049)      \\ \midrule
    \multirow{4}{*}{Joint training} 
    & CLAM-MB \cite{lu2021data}              & 95.015 ($\pm$4.030)  & 0.621 ($\pm$0.109) & 0.619 ($\pm$0.107) & 0.624 ($\pm$0.110)      \\
    & TransMIL \cite{shao2021transmil}       & 94.609 ($\pm$1.791)  & 0.741 ($\pm$0.199) & 0.744 ($\pm$0.218) & 0.739 ($\pm$0.182)      \\
    & DSMIL \cite{dsmil}                     & 91.553 ($\pm$6.621)  & 0.714 ($\pm$0.137) & 0.719 ($\pm$0.145) & 0.717 ($\pm$0.139)      \\
    & DGCN \cite{dgcn}                       & 91.165 ($\pm$2.857)  & 0.679 ($\pm$0.022) & 0.681 ($\pm$0.026) & 0.681 ($\pm$0.020)      \\ \midrule
    Joint Training + $\mathcal{T}$ 
    & MECFormer (ours)   & \textcolor{red}{96.930 ($\pm$1.748)}  & \textcolor{red}{0.959 ($\pm$0.027)} & \textcolor{red}{0.970 ($\pm$0.029)}   & \textcolor{red}{0.951 ($\pm$0.032)}  \\
    \bottomrule
    \end{tabular}}
    \label{tab:main_uni}
    \end{table}

\begin{figure}[http]
\centerline{\includegraphics[width=1\linewidth]{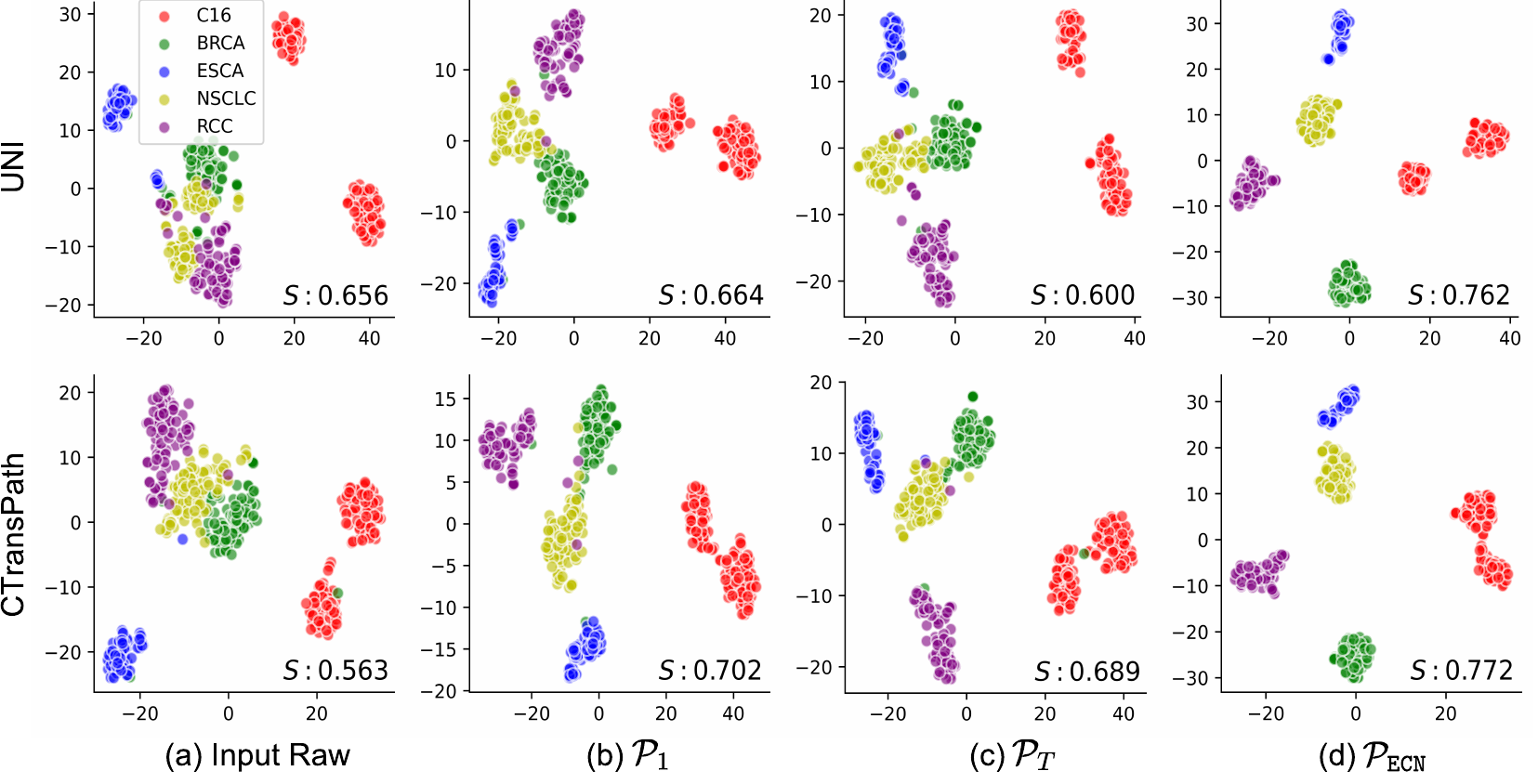}}
\caption{t-SNE visualizations for four scenarios: 1) Raw features extracted by off-the-shelf feature extractor $\mathcal{G}(\cdot)$, 2) $\mathcal{P}_{1}$, 3) $\mathcal{P}_{T}$, and 4) $\mathcal{P}_{\texttt{ECN}}$. $S$ denotes a Silhouette score.}
\label{fig:tnse}
\end{figure}

\subsection{Ablation Results} To deepen our understanding of MECFormer, we carried out ablation studies on two aspects: the effectiveness of 1) $\mathcal{P}_{\texttt{ECN}}$ and 2) the language decoder $\mathcal{D}$. Here, we report the overall Acc, F1, Recall, and Precision over the five tasks. The detailed results for each task are available in the Supplementary Material.

\subsubsection{Effectiveness of \texttt{ECN}.} To assess the effectiveness of \texttt{ECN}, we analyzed three types of projection layers: 1) $\mathcal{P}_{1}$: a single projection layer, i.e., a MLP, for all tasks, 2) $\mathcal{P}_{T}$: a single projection layer per task, i.e., $ T $ MLPs, and 3) $\mathcal{P}_{\texttt{ECN}}$. 
Using each of the three projection layers, we projected the patch embeddings $\mathbf{v}^{(0)} \in \mathbb{R}^{N \times d_{model}}$, computed the mean value per channel, and obtained one representative embedding vector $ {v}^{(0)} \in \mathbb{R}^{d_{model}} $ for each WSI. Then, t-SNE is applied to visualize the distribution of WSIs. We also computed the Silhouette score for each task. Figure \ref{fig:tnse} illustrates the visualization results and Silhouette scores.
It is obvious that the embedding vectors produced by $\mathcal{P}_{\texttt{ECN}}$ well separate differing tasks without any overlap between them, which is confirmed by the highest Silhouette scores of 0.762 and 0.772 for UNI and CTransPath, respectively.
As for other models, there were overlaps between tasks, which can explain the generation of out-of-distribution categories. 

Table \ref{tab:ablation_projection} shows the classification results by using each of the three projection layers. Using $\mathcal{P}_{1}$ and $\mathcal{P}_{T}$, the performance was considerably dropped compared to $\mathcal{P}_{\texttt{ECN}}$; for example, using CTransPath as $\mathcal{G}$, Acc, F1, Recall, and Precision decreased by $\ge$3.149\% Acc, $\ge$0.031 F1, $\ge$0.029 Recall, and $\ge$0.028 Precision, respectively; employing UNI as $\mathcal{G}$, a performance drop of $\ge$2.749\%, $\ge$0.032, $\ge$0.028, and $\ge$0.033 was observed for Acc, F1, Recall, and Precision, respectively. 
These results validate the importance of $\texttt{ECN}$ in the design of MECFormer.

\begin{table}[]
\centering
\caption{Ablation results on the three projection layer: $\mathcal{P}_{1}$, $\mathcal{P}_{T}$, and $\mathcal{P}_{\texttt{ECN}}$.}
\resizebox{1\textwidth}{!}{%
\begin{tabular}{P{2cm}|P{2cm}|P{3cm}|P{3cm}|P{3cm}|P{3cm}}
\toprule
$\mathcal{G}$                           & \textbf{Type}                     & \textbf{Acc (\%)}          & \textbf{F1}            & \textbf{Recall}        & \textbf{Precision}     \\ \midrule
\multirow{3}{*}{CTransPath} 
& $\mathcal{P}_1$ & 90.438 ($\pm$0.038) & 0.904 ($\pm$0.007) & 0.904 ($\pm$0.014) & 0.914 ($\pm$0.017) \\
& $\mathcal{P}_{T}$       & 91.259 ($\pm$2.131)            & 0.905 ($\pm$0.028)            &  0.901 ($\pm$0.035)            &  (0.908$\pm$0.024)            \\
& $\mathcal{P}_{\texttt{ECN}}$ & \textbf{94.408 ($\pm$1.879)} & \textbf{0.936 ($\pm$0.027)} & \textbf{0.933 ($\pm$0.031)} & \textbf{0.942 ($\pm$0.022)} \\ \midrule
\multirow{3}{*}{UNI}        
& $\mathcal{P}_1$ & 93.118 ($\pm$2.513) & 0.918 ($\pm$0.031) & 0.926 ($\pm$0.031) & 0.916 ($\pm$0.031) \\
& $\mathcal{P}_{T}$ & 92.410 ($\pm$3.092)            & 0.915 ($\pm$0.032)            & 0.922 ($\pm$0.040)            & 0.913 ($\pm$0.023)            \\
& $\mathcal{P}_{\texttt{ECN}}$ & \textbf{95.867 ($\pm$0.829)} & \textbf{0.950 ($\pm$0.011)} & \textbf{0.954 ($\pm$0.013)} & \textbf{0.949 ($\pm$0.008)} \\ \bottomrule
\end{tabular}}
\label{tab:ablation_projection}
\end{table}

\subsubsection{Effectiveness of $\mathcal{D}$.} To investigate the usefulness of the language decoder $\mathcal{D}$ in a unified model for multi-task learning, we repeated the \textit{joint learning} experiments on the five tasks without $\mathcal{D}$, i.e., the classifier conducts the classification for all categories across five tasks. The results are reported in Table \ref{tab:ablation_decoder}. Without $\mathcal{D}$, the classification performance was consistently dropped regardless of the evaluation metrics and the type of $\mathcal{G}$. Specifically, with CTransPath as $\mathcal{G}$, the performance drops were 2.215 Acc, 0.025 F1, 0.022 Recall, and 0.021 Precision; with UNI as $\mathcal{G}$, an Acc of 1.518, an F1 of 0.022, a Recall of 0.024, and a Precision of 0.019 decreased. These ablation results confirm the effectiveness of the language decoder $\mathcal{D}$ in our design, which facilitates flexible and adaptable classification and improves the overall performance.

\begin{table}[]
\centering
\caption{Ablation results on the language decoder $\mathcal{D}$.}
\resizebox{1\textwidth}{!}{%
\begin{tabular}{P{2cm}|P{3cm}|P{3cm}|P{3cm}|P{3cm}|P{3cm}}
\toprule
$\mathcal{G}$                           & \textbf{Method}                     & \textbf{Acc (\%)}          & \textbf{F1}            & \textbf{Recall}        & \textbf{Precision}     \\ \midrule
\multirow{2}{*}{CTransPath} & MECFormer \texttt{w/o} $\mathcal{D}$  & 92.193 ($\pm$2.478)             & 0.911 ($\pm$0.032)            & 0.911 ($\pm$0.024)            & 0.921 ($\pm$0.034)            \\
& MECFormer & \textbf{94.408 ($\pm$1.879)} & \textbf{0.936 ($\pm$0.027)} & \textbf{0.933 ($\pm$0.031)} & \textbf{0.942 ($\pm$0.022)} \\ \midrule
\multirow{2}{*}{UNI}        & MECFormer \texttt{w/o} $\mathcal{D}$  & 94.169 ($\pm$2.163)             & 0.928 ($\pm$0.030)            & 0.930 ($\pm$0.036)            & 0.930 ($\pm$0.025)            \\
& MECFormer  & \textbf{95.687 ($\pm$0.829)} & \textbf{0.950 ($\pm$0.011)} & \textbf{0.954 ($\pm$0.013)} & \textbf{0.949 ($\pm$0.008)} \\ \bottomrule
\end{tabular}}
\label{tab:ablation_decoder}
\end{table}

\section{Conclusion}

We propose MECFormer for a unified and multi-task WSI classification. MECFormer permits the predictions for multiple tasks across different organs using a single unified generative model, assuming that the target task is specified. To handle multiple tasks simultaneously, we introduce $\texttt{ECN}$, a projection layer that simulates expert consultation by gathering knowledge through effective gate weighting. For flexible classification, the prediction step is devised as an auto-regressive decoding process using a language decoder. Through extensive experiments, we have demonstrated the superior performance of MECFormer and validated its design for multi-task WSI classification. MECFormer undoubtedly has the potential to be extended to address multi-task learning challenges in other domains.

\section*{Acknowledgments}
This study was supported by a grant of the National Research Foundation of Korea (NRF) (No. 2021R1A2C2014557 and No. RS-2024-00397293) and Institute of Information \& communications Technology Planning \& Evaluation (IITP) (No. RS-2022-00167143).


%
%
\bibliographystyle{splncs04}
\bibliography{main}
\end{document}